\def\paperID{3936}
\def\confName{CVPR}
\def\confYear{2024}

\def\paperTitle{MOHO: Learning Single-view Hand-held Object Reconstruction with Multi-view Occlusion-Aware Supervision}

\def\authorBlock{
    Chenyangguang Zhang$^{1}$*,
    Guanlong Jiao$^{1}$*, 
    Yan Di$^{2}$, 
    Gu Wang$^{1}$,
    Ziqin Huang$^{1}$,\\
    Ruida Zhang$^{1}$,
    Fabian Manhardt$^{3}$,
    Bowen Fu$^{1}$,
    Federico Tombari$^{2,3}$
    and Xiangyang Ji$^{1}$\\
    \textsuperscript{1}Tsinghua University, \textsuperscript{2}Technical University of Munich,
    \textsuperscript{3} Google\\
    \tt\small{\{zcyg22@mails., xyji@\}tsinghua.edu.cn}
    \thanks{*Authors with equal contributions.}
    \thanks{Codes and datasets: \url{https://github.com/ZhangCYG/MOHO}}
}

\newif\ifreview 
\newif\ifarxiv 
\newif\ifcamera \newcommand{\cameraready}{\cameratrue}
\newif\ifrebuttal 

\cameraready 

\pdfoutput=1
\documentclass[10pt,twocolumn,letterpaper]{article}
\ifreview \usepackage[review]{cvpr} \fi
\ifarxiv \usepackage[pagenumbers]{cvpr} \fi
\ifrebuttal \usepackage[rebuttal]{cvpr} \fi
\ifcamera \usepackage{cvpr} \fi

\usepackage{graphicx}	
\usepackage{amsmath}	
\usepackage{amssymb}	
\usepackage{booktabs}
\usepackage{times}
\usepackage{microtype}
\usepackage{epsfig}
\usepackage[table,xcdraw,dvipsnames]{xcolor}
\usepackage{caption}
\usepackage{float}
\usepackage{placeins}
\usepackage{color, colortbl}
\usepackage{stfloats}
\usepackage{enumitem}
\usepackage{tabularx}
\usepackage{xstring}
\usepackage{multirow}
\usepackage{xspace}
\usepackage{url}
\usepackage{subcaption}
\usepackage{xcolor}
\usepackage{pifont}
\usepackage[hang,flushmargin]{footmisc}

\ifcamera \usepackage[accsupp]{axessibility} \fi

% ...

\newcommand{\nbf}[1]{{\noindent \textbf{#1.}}}

\newcommand{\supp}{supplementary material\xspace}
\ifarxiv \renewcommand{\supp}{appendix\xspace} \fi

\newcommand{\R}[1]{{%
    \textbf{%
        \ifstrequal{#1}{1}{\textcolor{red}{R#1}}{%
        \ifstrequal{#1}{2}{\textcolor{blue}{R#1}}{%
        \ifstrequal{#1}{3}{\textcolor{magenta}{R#1}}{%
        \ifstrequal{#1}{4}{\textcolor{teal}{R#1}}{%
                           \textcolor{cyan}{R#1}%
        }}}}%
    }%
}}

\newlength\savewidth\newcommand\shline{\noalign{\global\savewidth\arrayrulewidth
  \global\arrayrulewidth 1pt}\hline\noalign{\global\arrayrulewidth\savewidth}}

\newcommand{\tablestyle}[2]{\setlength{\tabcolsep}{#1}\renewcommand{\arraystretch}{#2}\centering\footnotesize}  

\newcommand{\cmark}{\ding{51}}
\newcommand{\xmark}{\ding{55}}  

\usepackage{xr-hyper}

\makeatletter
\newcommand*{\addFileDependency}[1]{
  \typeout{(#1)}
  \@addtofilelist{#1}
  \IfFileExists{#1}{}{\typeout{No file #1.}}
}

\makeatother

\definecolor{cvprblue}{rgb}{0.21,0.49,0.74}
\usepackage[pagebackref,breaklinks,colorlinks,citecolor=cvprblue]{hyperref}
\usepackage[capitalize]{cleveref}
\crefname{section}{Sec.}{Secs.}
\crefname{table}{Table}{Tables}
\crefname{figure}{Fig.}{Figs.}

\begin{document}
\title{\paperTitle}
\author{\authorBlock}
\maketitle


\begin{abstract}

Previous works concerning single-view hand-held object reconstruction typically rely on supervision from 3D ground-truth models, which are hard to collect in real world.
In contrast, readily accessible hand-object videos offer a promising training data source, but they only give heavily occluded object observations.
In this paper, we present a novel synthetic-to-real framework to exploit \textbf{M}ulti-view \textbf{O}cclusion-aware supervision from hand-object videos for \textbf{H}and-held \textbf{O}bject reconstruction (MOHO) from a single image, tackling two predominant challenges in such setting: hand-induced occlusion and object's self-occlusion.
First, in the synthetic pre-training stage, we render a large-scaled synthetic dataset SOMVideo with hand-object images and multi-view occlusion-free supervisions, adopted to address hand-induced occlusion in both 2D and 3D spaces.
Second, in the real-world finetuning stage, MOHO leverages the amodal-mask-weighted geometric supervision to mitigate the unfaithful guidance caused by the hand-occluded supervising views in real world.
Moreover, domain-consistent occlusion-aware features are amalgamated in MOHO to resist object's self-occlusion for inferring the complete object shape.
Extensive experiments on HO3D and DexYCB datasets demonstrate 2D-supervised MOHO gains superior results against 3D-supervised methods by a large margin.

\end{abstract}
\section{Introduction}

\begin{figure}[t]
    \centering
    \includegraphics[width=0.50\textwidth]{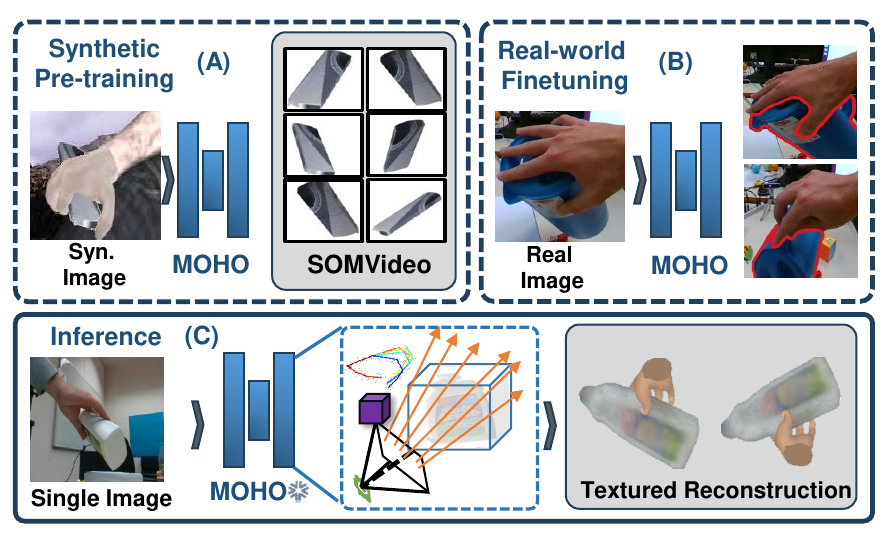}
    \vspace{-0.6cm}
    \caption{
    As a synthetic-to-real framework, MOHO is pre-trained by the rendered occlusion-free supervisions on SOMVideo, and then finetuned by the real-world hand-occluded supervising views.
    In the inference stage, MOHO generates the photorealistic reconstructed mesh given a single reference view, resisting both hand-induced occlusion and object's self-occlusion.}
    \label{fig:teaser}
\vspace{-0.7cm}
\end{figure}

Understanding hand-object interaction is becoming increasingly important in many practical scenarios including robotics \cite{zheng2023cams,li2020hoi}, augmented and virtual reality \cite{leng2021stable}, as well as embodied artificial intelligence systems \cite{xu2023unidexgrasp,li2023behavior}.
Although there exist previous works \cite{chao2021dexycb,yang2022oakink,huang2022reconstructing,ye2023diffusion} aiming at reconstructing fine hand-held object meshes from multi-view image sequences, single-view methods \cite{ye2022s,chen2022alignsdf,chen2023gsdf, hasson2019learning, karunratanakul2020grasping,zhang2023ddf} are drawing more attention recently since they can be applied more conveniently in real-world environment.

Given the ill-posed nature of single-view reconstruction, current top-preforming methods \cite{ye2022s,chen2022alignsdf,chen2023gsdf} typically use Signed Distance Fields (SDFs) as the geometric representation and employ 3D ground-truth meshes as supervision for training.
However, the applicability of such approaches in real-world scenarios is highly challenging, as obtaining clean and precise object meshes remains a formidable task.
In contrast, readily accessible raw videos capturing hands interacting with objects offer a promising training data source. 
Nevertheless, leveraging these videos as multi-view supervision for single-view hand-held object reconstruction introduces two significant challenges: hand-induced occlusion and object's self-occlusion. 
Firstly, hand-induced occlusion is an unavoidable issue in our easily obtained training data, leading to frequent instances of incomplete object views as objects are manipulated by hands. 
This incompleteness poses a significant hurdle for the network in effectively learning the reconstruction of the complete object shape. 
Thus, we adopt additional occlusion-free information from synthetic environments to mitigate the unfaithful guidance caused by the occluded supervising views in real world.
Additionally, the single-view setting exacerbates the problem with object's self-occlusion, as only one reference view is available, leaving the visible portion of the object incomplete and further complicating the task of enabling the network to recover the object's full shape.
Therefore, the occlusion-aware features need to be imposed for the network for full reconstruction.

To address the aforementioned problems, we present a novel synthetic-to-real framework to exploit \textbf{M}ulti-view \textbf{O}cclusion-aware supervision from hand-object videos for single-view \textbf{H}and-held \textbf{O}bject reconstruction (MOHO).
First, in the synthetic pre-training stage, we render a large-scale synthetic dataset SOMVideo with hand-object images and multi-view occlusion-free supervisions (\cref{fig:teaser} (A)).
MOHO takes one hand-object image as input and the other occlusion-free image describing the complete object in a novel view as supervision.
Thus, MOHO is empowered to remove hand-induced occlusion in 3D space.
Simultaneously, an auxiliary 2D amodal mask recovery head is integrated into the pre-training process, which predicts the hand-occluded parts of the object in the reference view.
Second, in the real-world finetuning stage, we freeze the 2D amodal mask recovery head to establish the amodal-mask-weighted geometric supervision, designed to combat the incomplete and defective supervisions presented by real-world hand-occluded videos (\cref{fig:teaser} (B)).
Moreover, in order to overcome object's self-occlusion in the whole synthetic-to-real process, we leverage domain-consistent occlusion-aware features including generic semantic cues and hand-articulated geometric embeddings.
These features are obtained with small cross-domain discrepancy, indicating which portions are visible in the reference view as well as hallucinating the shape of the self-occluded object surfaces.
Consequently, MOHO recovers complete 3D shape with photorealistic textures of the hand-held object via the geometric volume rendering technique during real-world inference (\cref{fig:teaser} (C)).

To summarize, our main contributions are threefold:
\begin{itemize}
    \item We propose a synthetic-to-real framework MOHO to pursue photorealistic hand-held object reconstruction from a single-view image without relying on 3D ground-truth supervision.
    To mitigate hand-induced occlusion, the rendered SOMVideo is adopted in the synthetic pre-training stage for occlusion-free supervisions, while the amodal-mask-weighted geometric supervision is proposed during the real-world finetuning. 
    \item 
    The domain-consistent occlusion-aware features are exploited in order to overcome object's self-occlusion in the whole synthetic-to-real process.
    \item 
    Extensive experiments on real-world datasets HO3D~\cite{hampali2020honnotate} and DexYCB~\cite{chao2021dexycb} demonstrate that 2D-supervised MOHO gains superior results against 3D-supervised methods.
\end{itemize}
\section{Related Works}
\noindent\textbf{Hand and Object Pose Estimation.}
The separate regression of hand pose and object pose constitutes a methodological stream for reconstructing hand-held objects.
Hand pose estimation from RGB(-D) input can be broadly categorized into two streams: model-free methods that lift detected 2D keypoints to 3D joint positions~\cite{iqbal2018hand, mueller2018ganerated, mueller2019real, panteleris2018using, rogez20143d, rogez2015understanding, zimmermann2017learning}, as well as model-based approaches that estimate statistical models with low-dimensional parameters~\cite{boukhayma20193d, rong2020frankmocap, sridhar2015fast, zhang2019end, zhou2020monocular, romero2022embodied}.
On the other hand, many works focus on initially regressing object poses based on predefined object templates~\cite{wang2021gdr,di2021so,wang2019normalized,tian2020shape, di2022gpv}, and subsequently, they proceed to reconstruct object meshes. 
In contrast, MOHO stands apart by its capability to reconstruct agnostic objects without relying on any prior assumptions. 
Our approach adopts the MANO model~\cite{romero2022embodied}, which shows more robustness to occlusion~\cite{ye2022s}, to provide hand articulations.

\begin{figure*}[t]
    \centering
    \includegraphics[width=0.99\textwidth]{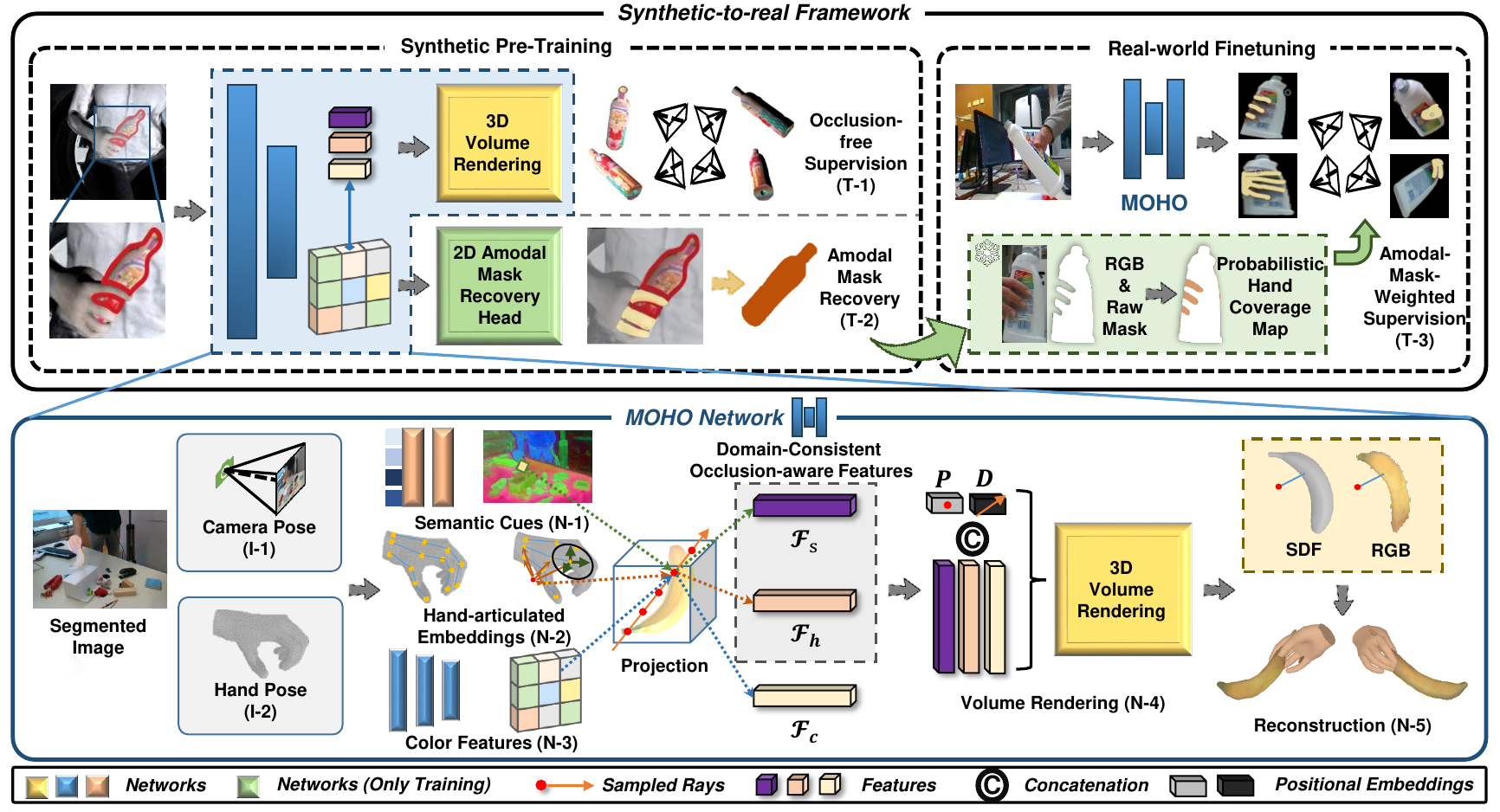}
     \caption{\textbf{Overview of MOHO.}
     \textbf{Synthetic-to-real Framework}: 
     We pre-train MOHO on the SOMVideo to resist hand-induced occlusion in both 3D (T-1) and 2D (T-2) spaces.
     The 2D recovered amodal masks are transferred into the real-world finetuning for releasing the incomplete hand-occluded supervisions (T-3).
     \textbf{Network}: 
     Given a segmented hand-object image as input, the estimated camera pose (I-1) and hand pose (I-2) are initialized by an off-line system \cite{rong2020frankmocap}.
     Subsequently, MOHO extracts domain-consistent occlusion-aware features including generic semantic cues (N-1) and hand-articulated geometric embeddings (N-2), as well as color features (N-3) for the volume rendering heads (N-4) to yield the textured mesh reconstruction of the full hand-held object (N-5).
     }
    \label{fig:pipeline}
\vspace{-0.5cm}
\end{figure*}

\noindent\textbf{Hand-held Object Reconstruction.}
Hand-held object reconstruction plays a crucial role in advanced understanding of human-object interaction. 
Previous works~\cite{garcia2018first, hampali2020honnotate, sridhar2016real, tekin2019h+} typically assume access to predefined object templates, employing joint regression techniques to estimate both hand poses and 6DoF object poses.
Some studies~\cite{chen2021joint, gkioxari2018detecting, liu2021semi, shan2020understanding} explore implicit feature fusion, incorporating geometric constraints~\cite{brahmbhatt2020contactpose, cao2021reconstructing, corona2020ganhand, grady2021contactopt, zhang2020perceiving} or promoting physical realism~\cite{tzionas2016capturing, pham2017hand} for such joint reasoning. 
Recent studies have shifted their focus toward directly reconstructing hand-held object meshes from monocular RGB inputs without relying on any prior assumptions.
For instance, \cite{hasson2019learning} develops a joint network that predicts object mesh vertices and MANO parameters of the hand, while \cite{karunratanakul2020grasping} predicts these parameters within a latent space. 
Additionally,  \cite{chen2022alignsdf,chen2023gsdf,ye2022s,zhang2023ddf} leverage Signed or Directed Distance Field (S/DDF) representations for hand and object shapes.
However, these methods require hard-to-collect 3D ground-truth data for training, limiting their applicability in real-world scenarios. 
Contrarily, MOHO alleviates the need for 3D ground-truth by exclusively utilizing 2D video supervision. 

\noindent\textbf{Volume Rendering Techniques.}
There has been a surge in the utilization of volume rendering techniques in the context of neural radiance fields (NeRFs)~\cite{mildenhall2021nerf,anciukevivcius2023renderdiffusion,li2023neuralangelo,wang2021neus,pavllo2023shape}, which have proven to be influential for advancing novel view synthesis and photorealistic scene reconstruction.
In the earlier stages, the focus is primarily on the development of scene-specific volume rendering~\cite{mildenhall2021nerf,wang2021neus}, where one network can only represent a single scene. 
Subsequent studies~\cite{jang2021codenerf,yu2021pixelnerf,xu2022sinnerf,lin2023vision,chen2023single} extend the scope of the problem to scene-agnostic, focusing on reconstruction of various objects from one single reference view or sparse views, which closely resembles the setting of single-view hand-held object reconstruction.
However, the performance of these previous methods is largely contingent on ideal conditions, where occlusion is not a significant factor. 
MOHO addresses this limitation by introducing a novel synthetic-to-real framework and leveraging domain-consistent occlusion-aware features. 
These additions aim to effectively mitigate the challenges posed by heavy occlusion in real world.

\section{Method}

\subsection{Overview}

As is shown in \cref{fig:pipeline}, MOHO, following the synthetic-to-real paradigm, is first pre-trained by the large-scaled rendered dataset SOMVideo for gaining the removal ability against hand-induced occlusion in both 3D and 2D spaces ((T-1) and (T-2)), and then finetuned on real-world videos with incomplete object observations.
We adopt the proposed amodal-mask-weighted geometric supervision (T-3) to mitigate the misguidance caused by heavily occluded real-world supervisions in the finetuning stage.
Meanwhile, the domain-consistent occlusion-aware features are leveraged in the whole synthetic-to-real process, which include generic semantic cues $\mathcal{F}_s$ extracted by the pre-trained DINO \cite{amir2022effectiveness} model $\mathcal{D}$ (N-1), and hand-articulated geometric embeddings $\mathcal{F}_h$ calculated by the predicted hand pose $\theta_A$ (N-2).
These domain-consistent occlusion-aware features, as well as color features $\mathcal{F}_c$ yielded by the CNN-based encoder $\phi$ (N-3), are concatenated as the condition for geometric volume rendering heads~\cite{wang2021neus} $\psi_S$, $\psi_C$ (N-4) to respectively predict the SDF value and the color density, enabling reconstruction of the agnostic occluded hand-held object without any instance priors.
During inference, given a single input reference image $\mathcal{I}$ depicting hand-object interaction, its corresponding camera pose $\mathcal{P}_I$ (I-1), offline-estimated object segmentation $\mathcal{S}_o$, and hand pose prediction $\theta_A$ (I-2), MOHO synthesizes the novel views as well as reconstructs the textured mesh of the complete hand-held object (N-5).

\subsection{Domain-consistent Occlusion-aware Features for Geometric Volume Rendering}
\label{cues}

Generating novel views and the full mesh for a hand-held object given by the only reference view is typically an ill-posed problem due to severe object's self-occlusion.
As \cite{xu2022sinnerf} demonstrates, reconstructing the whole scene from a single reference view may cause the volume rendering technique to generate unsatisfactory results, whose surface toward the reference view is recovered decently, but the reconstruction of unseen parts is degraded.
Hence, we need to feed sufficient information to the network for completing the unobserved area in the reference image.
Such imposed information should have least domain discrepancy, to ensure effective knowledge transfer in the whole synthetic-to-real framework.
Specifically, we exploit the domain-consistent occlusion-aware features from two aspects: generic semantic cues $\mathcal{F}_s$ (\cref{fig:pipeline} (N-1)) and hand-articulated geometric embeddings $\mathcal{F}_h$ (\cref{fig:pipeline} (N-2)).


\nbf{Generic Semantic Cues}
The generic semantic cues $\mathcal{F}_s$ are exploited to provide MOHO with high-level structural priors for amodal object perception.
Concretely, we harness semantic cues from the pre-trained DINO~\cite{amir2022effectiveness} model, which provides local descriptors with consistent structural information to indicate the position of the observed object parts within the whole shape.
Note that DINO is well demonstrated for its semantic stability across different domains~\cite{amir2022effectiveness}.
With such domain-consistent semantic cues, MOHO learns to complement the full object better from the partial observation under the whole synthetic-to-real process. 
Specifically, the pre-trained DINO model $\mathcal{D}$ extracts the patch-wise feature maps $\mathcal{F}^I_s = \mathcal{D}(\mathcal{I})\odot\mathcal{S}_o$.
We use the top three principal components of $\mathcal{F}^I_s$ by principal component analysis (PCA) considering the trade-off between efficiency and performance.
Since MOHO adopts the volume rendering technique, the feature map $\mathcal{F}^I_s$ needs to be converted to the features corresponding to the 3D sampled points.
Given the camera poses $\mathcal{P}_I$ and camera parameters $\mathcal{K}$, the 3D points $\{P_i\}_{i=1}^n$ along the sampled rays are first projected onto the image plane to get the corresponding pixel positions.
The patched color features of each sampled point $\mathcal{F}^i_s$ are fetched on $\mathcal{F}^I_s$ via bilinear interpolation.

\nbf{Hand-articulated Geometric Embeddings}
Considering that the holding hand shape implies the unobserved hand-held object shape, we add the hand-articulated geometric embeddings $\mathcal{F}^i_h$ for each sampled point $P_i$.
The adopted embeddings are explicitly yielded by calculating the geodesic distances from the sampled point $P_i$ to the nearest hand joints.
Such explicit embeddings remain stable and consistent during the whole synthetic-to-real transferring without any domain gaps.  
Specifically, we first use an offline hand pose estimator \cite{rong2020frankmocap} to get $\theta_A$ from the reference image. 
Then, we run forward kinematics of MANO model~\cite{romero2022embodied} to derive the transformation $T(\theta_A)$ as well as the hand joint coordinates.
Afterwards, the sampled point $P_i$ is mapped to the nearest $K$ hand joint coordinates by their transformation matrices. 
Finally, such $K$ positions of the sampled point in the nearest $K$ hand joint coordinates are concatenated as $\mathcal{F}^i_h$ to provide the distance information.
We select $K=6$ during the implementation. 
Notably, we treat the hand articulation locally using nearest hand joints rather than globally using all hand joints as the previous methods \cite{ye2022s,chen2023gsdf} do.
We find that taking all the joints' coordinates is unnecessary and leads to more complexity empirically (\cref{ablation}). 

\nbf{Color Features}
Since MOHO needs to recover the texture of the hand-held object, we follow \cite{yu2021pixelnerf} to use a ResNet34~\cite{he2016deep} as the image encoder $\phi$ to extract the object color feature map by $\mathcal{F}^I_c = \phi(\mathcal{I}\odot\mathcal{S}_o)$ of the reference view $\mathcal{I}$.
The $\mathcal{F}^i_c$ of the sampled 3D point $P_i$ is obtained by the same projection and interpolation operations as the generation of $\mathcal{F}^i_s$.

\nbf{Conditional Geometric Volume Rendering}
After incorporating all the domain-consistent occlusion-aware features, we construct the conditional geometric volume rendering technique to render novel views as well as generate textured meshes (\cref{fig:pipeline} (N-4)).
Specifically, given 3D sampled points $\{P_i\}$, ray directions $\{D_i\}$ and corresponding point features $\{\mathcal{F}^i_{con}\} = \{\text{Cat}(\mathcal{F}^i_s, \mathcal{F}^i_h, \mathcal{F}^i_c)\}$ extracted from the single reference view $\mathcal{I}$, a geometric field $\psi_S$ predicting the SDF value $s_i = \psi_S(P_i | \mathcal{F}^i_{con})$ as well as a color field $\psi_C$ predicting the RGB density $c_i = \psi_C(P_i, D_i | \mathcal{F}^i_c)$ are constructed. 
As for volume rendering, 3D points $\{P_i\}$ are sampled along camera rays by $\{P\}$ = $\{P(z) | P(z) = O + zD, z \in [z_n, z_f]\}$, where $O$ denotes the origin of the camera, $D$ refers to the viewing direction of each pixel, $z_n$, $z_f$ are the near and far bounds of the ray.
$O$ and $D$ are calculated by the input camera pose $\mathcal{P}_I$ and camera intrinsic $\mathcal{K}$.
Then, the color of the pixel is rendered by
\begin{equation}
\hat{c} = \int_{z_n}^{z_f} \omega(z) \psi_C(P(z), D | \mathcal{F}_c) dz,
\end{equation}
where $\omega(z) = T(z) \rho(z)$ is an unbiased and occlusion-aware function proposed by \cite{wang2021neus}, converting the SDF value $\psi_S(P | \mathcal{F}_{con})$ to $T(z), \rho(z)$ by $T(z) = \exp\left(-\int_{z_n}^{z_f}\rho(z)dz\right)$, $\rho(z) = \max\left(\frac{-\frac{d \sigma^h}{dz}\left(\psi_S(P(z) | \mathcal{F}_{con})\right)}{\sigma^h\left(\psi_S(P(z) | \mathcal{F}_{con})\right)},0\right)$.
$\sigma^h$ here denotes the Sigmoid function with a trainable parameter $h$.
During implementation, the formulations above are numerically discretized as referred to \cite{wang2021neus}.

\subsection{Synthetic-to-real Training Framework}
\label{training}

We propose a synthetic-to-real training framework for MOHO to overcome the omnipresent hand-induced occlusion met in real-world single-view hand-held object reconstruction.
In the synthetic pre-training stage, we foster MOHO for the capability to be aware of the hand-occluded regions of the object in both 3D and 2D spaces, with the utilization of our large-scaled rendered dataset SOMVideo.
For removing hand-induced occlusion in 3D space, MOHO inputs the hand-occluded reference view $\mathcal{I}\odot\mathcal{S}_o$ ($\odot$ means bitwise multiplication), and is supervised by the synthetic complete object in novel views (\cref{fig:pipeline} (T-1)).
Further, an auxiliary 2D amodal mask recovery head $\Gamma$ (\cref{fig:pipeline} (T-2)) is utilized to predict the probabilistic hand coverage map in 2D space.
After pre-training, MOHO is finetuned with real-world hand-object videos, so as to be better applied for real-world inference.
However, real-world hand-object multi-view images oftentimes contain truncated regions and incomplete views, resulting in detrimental effects when directly used for training.
Naively utilizing the defective masks of the hand-occluded objects misleads the network to reconstruct patchy geometric surfaces.
Thus, the predicted hand coverage maps from the pre-trained 2D head are leveraged on real-world data (\cref{fig:pipeline} (T-3)) to construct a soft constraint and introduce the amodal-mask-weighted geometric supervision for reconstructing full hand-held objects. 

\begin{figure}
    \centering
    \includegraphics[width=0.98\linewidth]{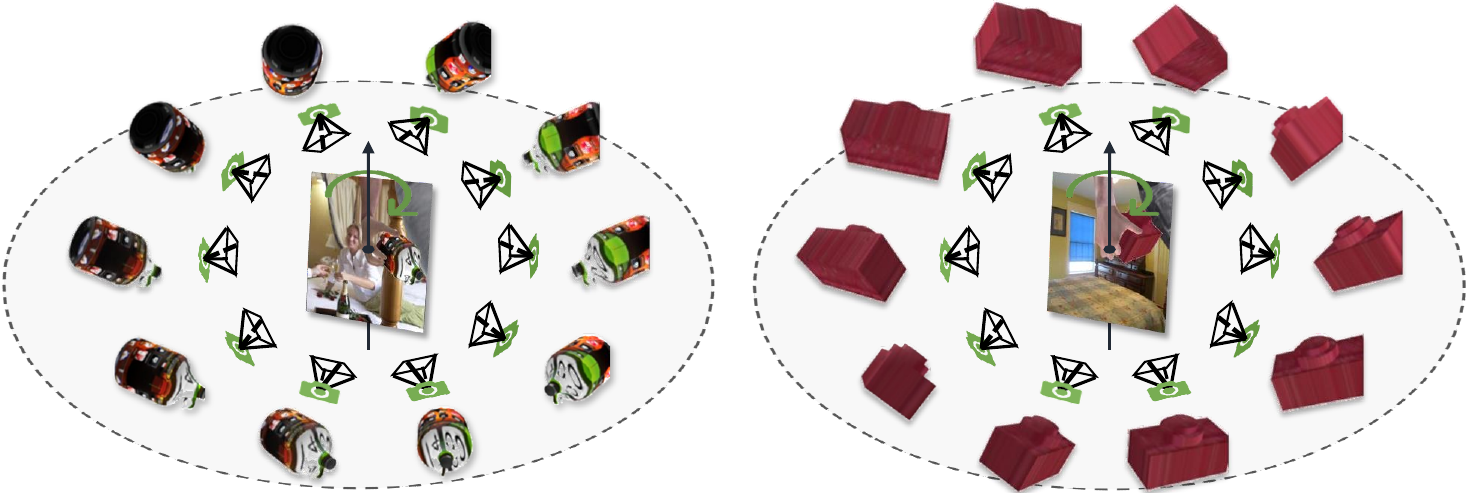}
    \vspace{-0.1cm}
    \caption{Visual illustration of SOMVideo rendered with occlusion-free multi-view supervisions.}
    \vspace{-0.5cm}
\label{fig:SOMVideo}
\end{figure}

\nbf{Synthetic Object Manipulation Video (SOMVideo) Dataset}
Current video datasets capturing hand-object interactions \cite{hampali2020honnotate,chao2021dexycb} are collected in real world.
They typically contain limited object instances and are unfriendly for constructing occlusion-free supervisions for purely 2D-supervised methods.
Therefore, following the generation pipeline of synthetic object manipulation scenes \cite{hasson2019learning}, we render the \textbf{S}ynthetic \textbf{O}bject \textbf{M}anipulation Video (SOMVideo) dataset for MOHO, boasting large-scaled hand-object images as well as corresponding occlusion-free multi-view supervisions (Fig. \ref{fig:SOMVideo}).

Generally, SOMVideo is synthesized by setting up the hand-object interaction scene and moving the camera.
First, we select the same 2,772 objects as the ObMan \cite{hasson2019learning} dataset from ShapeNet~\cite{chang2015shapenet}.
Objects' textures are randomly sampled from the texture maps provided by ShapeNet.
Then, the grasping generation procedure adopting GraspIt~\cite{miller2004graspit} and the body and hand models in the SMPL+H ~\cite{loper2015smpl,romero2022embodied} are used for the hand-object interaction scene setup.
Finally, we add rotations and translations on the camera towards the hand-object interaction scene to capture hand-object multi-view images.
Besides, to enable occlusion-free supervisions, we render corresponding video clips containing only objects without hands and bodies by setting the scene without SMPL+H models and keeping the camera parameters the same.
SOMVideo consists of 141,550 scenes in total, in which each hand-object scene is captured by 10 views.
Each corresponding occlusion-free video clip for supervision is also captured from 10 same view angles.
For more rendering details, please refer to the \supp.

\nbf{Synthetic Pre-Training with SOMVideo}
MOHO is first pre-trained to remove hand-induced occlusion in 3D space by the rendered occlusion-free supervisions.
Concretely, the input reference view fed to MOHO is a hand-object image, while the supervision views are corresponding rendered occlusion-free pictures from novel poses (\cref{fig:pipeline} (T-1)).
During each iteration in the pre-training, one hand-object reference image is fed to MOHO, while 8 novel views sampled from the corresponding occlusion-free video are regarded as supervision.

Simultaneously, an auxiliary 2D amodal mask recovery head $\Gamma$ is utilized in the pre-training.
The 2D recovery head, whose architecture refers to \cite{chen2018encoder}, predicts the hand-covered object's parts $\hat{M}_I^{ho}$ in the input reference view $\mathcal{I}$ by $\hat{M}_I^{ho} = \Gamma(\mathcal{F}^I_c)$ (\cref{fig:pipeline} (T-2)), where $\mathcal{F}^I_c$ is the color feature maps defined in \cref{cues}.
The supervision of this head is enforced by the binary cross-entropy loss between $\hat{M}_I^{ho}$ and $M_I^{co} \ominus M_I$, where $M_I^{co}$ means the rendered complete mask of the input reference view $\mathcal{I}$, $M_I$ refers to the input hand-occluded object mask, and $\ominus$ means bitwise subtraction.
The benefits of incorporating such 2D perception are twofold.
First, 2D hand coverage perception strengthens the ability of MOHO to handle hand-induced occlusion patterns.
Second, considering more cross-domain consistency of the 2D neural network \cite{cardace2023exploiting}, we exploit the predictions of this 2D head for the real-world finetuning stage to promote the knowledge transfer about hand-induced occlusion removal learned in the pre-training stage.
To this end, we freeze the 2D recovery head during the real-world finetuning stage, and infer the probabilistic hand coverage maps (\cref{fig:pipeline} (T-3)).
These maps are regarded as the relaxed constraints for the proposed amodal-mask-weighted geometric supervision.

\nbf{Real-World Finetuning with Amodal-Mask-Weighted Supervision}
After pre-training, MOHO is finetuned on hand-object videos from real-world datasets which typically suffer from partial observations caused by hand-induced occlusion.
Therefore, we introduce the amodal-mask-weighted geometric supervision, taking the probabilistic hand coverage maps predicted by the pre-trained 2D amodal mask recovery head in real world into consideration (\cref{fig:pipeline} (T-3)).
The amodal-mask-weighted loss is defined by
\begin{equation}
\mathcal{L}_{amw} = BCE(\hat{M}_T^{ho}\oplus M_T, \hat{O}_T),
\label{mw_loss}
\end{equation}
where $T$ refers to a novel target view, $\hat{M}_T^{ho}$ means the recovered amodal mask, $\hat{O}_T$ is the predicted object mask by the volume rendering heads and $\oplus$ means bitwise addition.
Having gained knowledge about how to handle hand occlusion with such supervision, MOHO is capable of inferring the shape of the complete object in real world.

\nbf{Volume Rendering Losses for Synthetic-to-real Training}
Several losses are designed for supervising the 3D volume rendering heads of MOHO to get geometric consistent surfaces as well as photorealistic texture results in the whole synthetic-to-real framework.
The overall loss function is defined as
\begin{equation}
\mathcal{L} = \mathcal{L}_{color} + \lambda_1 \mathcal{L}_{eik} + \lambda_2 \mathcal{L}_{mask} + \lambda_3 \mathcal{L}_{n_{ori}} + \lambda_4 \mathcal{L}_{n_{smo}}.
\label{total_loss}
\end{equation}
Thereby, the color loss $\mathcal{L}_{color}$ is derived by $\mathcal{L}_{color} = |\hat{C}_T - C_T|$, where $\hat{C}_T$ and $C_T$ mean the predicted and ground-truth color maps of a novel view $T$ respectively.
The Eikonal term~\cite{gropp2020implicit} $\mathcal{L}_{eik} = \frac{1}{n} \sum_{i} (||\nabla \psi_S(P_{i})||_2 - 1)^2$ is added for geometric regularization, in which $P_{i}$ refers to the sampled points for volume rendering and $n$ is the number of sampling points.
The mask loss $\mathcal{L}_{mask}$ is defined differently in the pre-training stage and the finetuning stage, for exploiting the occlusion-free supervisions of SOMVideo and adopting the amodal-mask-weighted supervision in real world respectively.
At the pre-training stage, the mask loss is defined as
\begin{equation}
\mathcal{L}_{mask} = BCE(M_T^{co}, \hat{O}_T),
\label{mask_loss}
\end{equation}
where $M_T^{co}$ refers to the occlusion-free object mask of a novel view $T$ in SOMVideo.
At the finetuning stage, the mask loss is substituted by $\mathcal{L}_{mask} = \mathcal{L}_{amw}$ defined in \cref{mw_loss}.
Two additional losses regularizing the predicted surface normals are used for restricting the orientation of visible normals towards the camera ($\mathcal{L}_{n_{ori}}$) \cite{verbin2022ref} and making the predictions smoother ($\mathcal{L}_{n_{smo}}$) \cite{sharma2021point}, which are detailed in the \supp.
The weighted factors are set to $\lambda_1 = 1.0$, $\lambda_2 = 1.0$, $\lambda_3 = 10^3$, $\lambda_4 = 10^{-2}$ during implementation.
All factors are kept the same in both the synthetic pre-training and real-world finetuning.

\section{Experiment}
\subsection{Experimental Setup}
\nbf{Datasets}
We conduct experiments on two representative real-world datasets capturing hand-object interactions, HO3D~\cite{hampali2020honnotate} and DexYCB~\cite{chao2021dexycb}.
HO3D~\cite{hampali2020honnotate} contains 77,558 images from 68 sequences with 10 different persons manipulating 10 different objects.
The pose annotations are yielded by multi-camera optimization pipelines.
We follow \cite{hampali2020honnotate} to split training and testing sets.
DexYCB \cite{chao2021dexycb} is currently one of the largest real-world hand-object video datasets \cite{chen2023gsdf}.
We follow \cite{chen2022alignsdf,yang2022artiboost} to concentrate on right-hand samples and use the official $s0$ split.
29,656 training samples and 5,928 testing samples are downsampled referring to the setting of \cite{chen2023gsdf}.
Note that for both datasets, MOHO only utilizes the RGB pictures, segmentations, and poses for training, but without any need for the 3D ground-truth meshes.

\nbf{Baselines}
3D-supervised baselines including Atlas-Net-based \cite{groueix2018papier} HO~\cite{hasson2019learning}, implicit-field-based GF~\cite{karunratanakul2020grasping}, SDF-based IHOI~\cite{ye2022s}, AlignSDF~\cite{chen2022alignsdf} and gSDF~\cite{chen2023gsdf} are adopted for geometric comparisons with MOHO.
We mainly compare the reconstructed meshes with them to demonstrate the ability of MOHO for surface reconstruction.
Moreover, several 2D-supervised object-agnostic NeRF-based baselines are implemented, including PixelNeRF~\cite{yu2021pixelnerf} and the more recent SSDNeRF~\cite{chen2023single}.
We follow their single-view reconstruction setting and use the same training data as MOHO.
We report both geometric reconstruction and novel view synthesis metrics against the NeRF-based baselines.

\begin{table}[t]
\begin{center}
\tablestyle{16pt}{1.1}
\begin{tabular}{@{}c|ccc@{}}
\shline
Method            & F-5 $\uparrow$      & F-10 $\uparrow$       & CD $\downarrow$  \\
\hline
HO \cite{hasson2019learning} &0.11  &0.22 &4.19  
\\
GF \cite{karunratanakul2020grasping} &0.12  &0.24  &4.96 
 \\
IHOI \cite{ye2022s} &0.28   &\textbf{0.50}   &1.53   \\
\hline
PixelNeRF \cite{yu2021pixelnerf}        &0.17 &0.32  &6.91\\
SSDNeRF \cite{chen2023single}        &0.25 &0.40  &2.60\\
Ours        &\textbf{0.31}  &\textbf{0.50}  &\textbf{0.91}\\
\shline
\end{tabular}
\end{center}
\vspace{-0.5cm}
\caption{Geometric results on HO3D \cite{hampali2020honnotate} compared with 3D supervised methods (top) and 2D supervised methods (bottom).
}
\label{tab:ho3d}
\vspace{-0.2cm}
\end{table}

\begin{table}[t]
\begin{center}
\tablestyle{14pt}{1.1}
\begin{tabular}{@{}c|ccc@{}}
\shline
Method            & F-5 $\uparrow$      & F-10 $\uparrow$       & CD $\downarrow$  \\
\hline
HO \cite{hasson2019learning} &0.38  &0.64 &0.42  
\\
GF \cite{karunratanakul2020grasping} &0.39  &0.66  &0.45
 \\
AlignSDF \cite{chen2022alignsdf} &0.41   &0.68   &0.39   \\
gSDF \cite{chen2023gsdf} &0.44   &0.71   &0.34   \\
\hline
PixelNeRF \cite{yu2021pixelnerf}        &0.25 &0.46  &0.94\\
SSDNeRF \cite{chen2023single}        &0.27 &0.49  &0.58\\
Ours w/o SYN        &0.52 &0.74  &0.18\\
Ours        &\textbf{0.60}  &\textbf{0.81}  &\textbf{0.15}\\
\shline
\end{tabular}
\end{center}
\vspace{-0.5cm}
\caption{Geometric results on DexYCB \cite{chao2021dexycb} compared with 3D supervised methods (top) and 2D supervised methods (bottom).
}
\label{tab:dexycb}
\vspace{-0.7cm}
\end{table}

\nbf{Evaluation Metrics}
For geometric metrics, we follow~\cite{chen2023gsdf,ye2022s} to uniformly sample 30,000 points on the reconstructed mesh, and report mean Chamfer Distance (CD, mm) and F-score at thresholds of 5mm (F-5) and 10mm (F-10).
For metrics of novel view synthesis, we randomly sample 10 images in each video as the input reference views and another 10 views as the target views for each input reference. 
We report average PSNR, SSIM \cite{wang2004image}, and LPIPS \cite{zhang2018unreasonable} of the whole video dataset. 
Only the region within the object mask is considered for the aim of object reconstruction. 

\nbf{Implementation Details}
We train MOHO on a single NVIDIA A100 GPU using an Adam optimizer with a learning rate of $10^{-3}$ for synthetic pre-training and $4\times 10^{-4}$ for real-world finetuning.
The learning rate is scheduled by the cosine decay to the minimum of $5\times 10^{-5}$.
In the pre-training, we randomly select one hand-object reference view as the network input and 8 occlusion-free target views for supervision at each iteration.
In the finetuning, the reference view and target views are selected in real-world video data.
We pre-train MOHO on SOMVideo for 300K iterations and the real-world finetuning stage continues for another 300K iterations.
For volume rendering, we use the same coarse-to-fine ray sampling technique as \cite{wang2021neus} by first uniformly sampling 40 points along the ray and then upsampling another 40 points near the coarsely predicted surface.
During training, we randomly sample 150 rays in the object bounding box for each picture, following the protocol of ray origin and direction sampling strategy of \cite{yu2021pixelnerf}.
Our SOMVideo data will be released along with our codes.
For more details about the network architecture and synthetic data generation, please refer to the \supp.

\begin{figure*}
    \centering
    \includegraphics[width=0.95\linewidth]{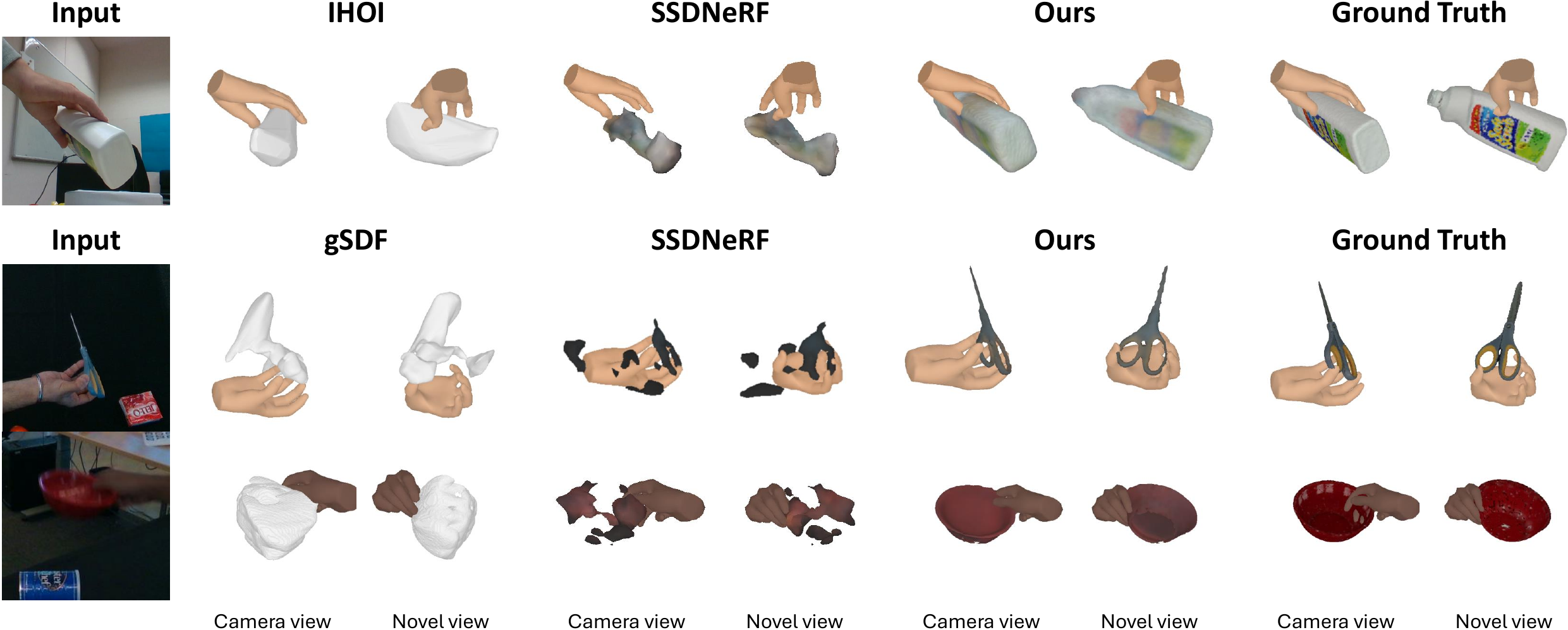}
    \vspace{-0.1cm}
    \caption{Visualization of textured meshes reconstructed by several baselines IHOI \cite{ye2022s}, gSDF \cite{chen2023gsdf}, SSDNeRF \cite{chen2023single} and MOHO on HO3D~\cite{hampali2020honnotate} (top) and DexYCB~\cite{chao2021dexycb} (bottom).
    The reconstruction results are exhibited on the camera view and one novel view.}
    \vspace{-0.2cm}
\label{fig:vis}
\end{figure*}

\subsection{Geometric Reconstruction}

We compare the quality of the geometric reconstruction ability of MOHO with two lines of methods including 3D-supervised baselines (typically SDF-based) and 2D-supervised baselines (typically object-agnostic NeRF-based).
\cref{tab:ho3d} and \cref{tab:dexycb} exhibit the geometric metrics on HO3D~\cite{hampali2020honnotate} and DexYCB~\cite{chao2021dexycb} respectively.
In addition, we analyze the \textbf{efficiency} of MOHO in the \supp.

In \cref{tab:ho3d}, we report all baselines' metrics following the setting of \cite{ye2022s} which utilizes the synthetic-to-real paradigm to release the problem of data scarcity and lack of diversity in real world.
All 3D-supervised methods in the top block strictly follow \cite{ye2022s} to initialize with models pre-trained on ObMan~\cite{hasson2019learning} and then finetune on HO3D~\cite{hampali2020honnotate}.
For 2D-supervised methods in the bottom block, we mimic the pre-training above using the hand-object multi-view images in SOMVideo for fair comparison.
Compared to 2D-supervised baselines, MOHO exceeds PixelNeRF by 82.3\% and the more recent SSDNeRF by 24.0\% on the F-5 metric.
This superiority demonstrates that due to the well-designed synthetic-to-real framework and the incorporation of the occlusion-aware features, MOHO better handles the extreme occlusion met when reconstructing hand-held objects from a single view.
In contrast, the previous object-agnostic NeRF-based methods are only illustrated to be effective under the ideal occlusion-free condition.
Compared with 3D-supervised methods, MOHO yields a significantly lower CD against the current top-performing approach IHOI by 40.5\%, while leads by more against previous HO and GF.
This indicates that the geometric surfaces reconstructed by MOHO contain much fewer outliers.

Experiments on the larger dataset DexYCB~\cite{chao2021dexycb} shown in \cref{tab:dexycb} further demonstrate the superiority of MOHO.
Due to the relatively sufficient data of DexYCB, the previous work~\cite{chen2023gsdf} reports metrics of prior 3D supervised baselines~\cite{chen2023gsdf,chen2022alignsdf,hasson2019learning,karunratanakul2020grasping} directly trained on the real-world data.
Thus, for fair comparison, we adopt this setting in \cref{tab:dexycb} for both 3D and 2D supervised baselines and also report metrics of MOHO without the synthetic pre-training stage using SOMVideo (Ours w/o SYN).
Thanks to the effective geometric volume rendering guided by the domain-consistent occlusion-aware features, MOHO w/o SYN achieves higher F-5 by 18.2\% than the state-of-the-art 3D-supervised gSDF, while obtaining 92.6\% superiority against the top-performing 2D supervised SSDNeRF on F-5 metric.
When considering our proposed synthetic-to-real training framework, the lead margin is further extended to 36.4\% and 122.2\% respectively.
This demonstrates that by considering the imposed features and training strategy together, MOHO is endowed with stronger robustness for handling various occlusion scenarios in real world.

\begin{table}[t]
\begin{center}
\tablestyle{2pt}{1.1}
\begin{tabular}{@{}c|ccc|ccc@{}}
\shline
\multirow{2}{*}{Method} & \multicolumn{3}{c|}{HO3D~\cite{hampali2020honnotate}} & \multicolumn{3}{c}{DexYCB~\cite{chao2021dexycb}} \\
& PSNR  $\uparrow$ & SSIM $\uparrow$ & LPIPS $\downarrow$ & PSNR $\uparrow$ & SSIM $\uparrow$ & LPIPS $\downarrow$ \\
\hline
PixelNeRF \cite{yu2021pixelnerf}        &24.82 &0.955  &0.055 &32.77 &0.986  &0.019 \\
SSDNeRF \cite{chen2023single}        &21.08 &0.943  &0.070      &32.83 &0.985  &0.022 \\
Ours        &\textbf{26.01}  &\textbf{0.960}  &\textbf{0.049} & \textbf{35.80}  &\textbf{0.989}  &\textbf{0.013} \\
\shline
\end{tabular}
\end{center}
\vspace{-0.5cm}
\caption{Novel view synthesis results.
}
\label{tab:nerf}
\vspace{-0.7cm}
\end{table}


\subsection{Novel View Synthesis}
As MOHO adopts the volume rendering technique, we also report its performance of novel view synthesis with other counterparts in \cref{tab:nerf}, to demonstrate the capability to recover the object texture from the single view input.
Notably, the previous predominant SDF-based methods \cite{ye2022s,chen2022alignsdf,chen2023gsdf,hasson2019learning,karunratanakul2020grasping} in the field cannot generate reconstruction results with object texture.
However, MOHO can not only get geometrically coherent object surfaces, but also yield photorealistic object texture, which enables it to adapt in more application scenarios.
Concretely, for novel view synthesis, MOHO leads by 4.8\% on PSNR and 10.9\% on LPIPS against PixelNeRF on HO3D and exceeds by 9.0\% and 40.9\% respectively compared with SSDNeRF on DexYCB, which illustrates the superior performance of MOHO than the NeRF-based competitors.

\nbf{Visualization}
To demonstrate both the abilities to get geometric surface as well as photorealistic texture, we visualize the textured meshes predicted by MOHO and compare them with 3D-supervised IHOI / gSDF and 2D-supervised SSDNeRF in \cref{fig:vis}.
The results show that the 2D-supervised baseline typically fails and only yields incomplete and patchy reconstructed meshes when given heavily occluded real-world supervision (Row 2, 3).
The 3D-supervised baselines obtain oversmoothed geometric surfaces without object texture.
Moreover, they fail to reconstruct tiny objects like the scissors in Row 2.
In contrast, MOHO is able to reconstruct geometrically coherent and photorealistic meshes.
More visualization results are shown in the \supp.

\subsection{Ablation Studies} \label{ablation}

\begin{table}
\begin{center}
\tablestyle{3pt}{1.1}
\begin{tabular}{@{}c c|cc|cc@{}}
\shline
\multirow{2}{*}{Method} & \multirow{2}{*}{Pre-training} & \multicolumn{2}{c|}{HO3D~\cite{hampali2020honnotate}} & \multicolumn{2}{c}{DexYCB~\cite{chao2021dexycb}} \\
& &F-5 $\uparrow$ &  CD $\downarrow$ &  F-5 $\uparrow$  &  CD $\downarrow$ \\
\hline
IHOI \cite{ye2022s}&\xmark &0.16  &2.06  &-  & -  \\
IHOI \cite{ye2022s}&\cmark &0.28  &1.53  &-  &-  \\
gSDF \cite{chen2023gsdf}& \xmark &-  & - &0.44  &0.34  \\
gSDF \cite{chen2023gsdf}&\cmark &-  & - &0.46  &0.30   \\
\hline
PixelNeRF \cite{yu2021pixelnerf}& \xmark &0.13  &14.07  &0.25  &0.94  \\
PixelNeRF \cite{yu2021pixelnerf}&\cmark &0.20  &6.02 &0.30  &0.54   \\
Ours & \xmark &0.22  &1.18  &0.52  &0.18   \\
Ours &\cmark~w/o AMW &0.29  &0.99  &0.57  &0.16  \\
Ours &\cmark &\textbf{0.31}  &\textbf{0.91}  &\textbf{0.60}  &\textbf{0.15}  \\
\shline
& & PSNR $\uparrow$      & LPIPS $\downarrow$   & PSNR $\uparrow$      & LPIPS $\downarrow$      \\
\hline
Ours & \xmark &24.96  &0.052  &35.41  &0.014   \\
Ours &\cmark &\textbf{26.01}  &\textbf{0.049}  &\textbf{35.80}  &\textbf{0.013}  \\
\shline
\end{tabular}
\end{center}
\vspace{-0.5cm}
\caption{Ablations for the synthetic-to-real training framework.}
\label{tab:training_ablation}
\vspace{-0.4cm}
\end{table}

\begin{table}[t]
\begin{center}
\tablestyle{8pt}{1.1}
\begin{tabular}{@{}cccc|ccc@{}}
\shline
$\mathcal{F}_s$-3 & $\mathcal{F}_s$-16   & $\mathcal{F}_h$-G  & $\mathcal{F}_h$-L &  F-5 $\uparrow$ &  F-10 $\uparrow$ &  CD $\downarrow$ \\
\hline
\xmark & \xmark & \xmark & \xmark&0.52  &0.72  &0.18  \\
\cmark & \xmark & \xmark & \xmark&0.54  &0.75  &0.17 \\
\xmark & \cmark & \xmark& \xmark &0.55  &0.76  &0.16\\
\cmark & \xmark & \cmark & \xmark&0.59 &0.79  &\textbf{0.15} \\
\cmark & \xmark & \xmark & \cmark&{\bf 0.60} &{\bf 0.81}  &\textbf{0.15} \\
\shline
\end{tabular}
\end{center}
\vspace{-0.5cm}
\caption{Ablations for the domain-consistent occlusion-aware features on DexYCB~\cite{chao2021dexycb}.}
\label{tab:feature_ablation}
\vspace{-0.7cm}
\end{table}

We conduct ablation studies from two main aspects, \ie, the effectiveness of the proposed synthetic-to-real training scheme (\cref{tab:training_ablation}), and the generic semantic cues as well as the hand-articulated geometric embeddings included in the domain-consistent occlusion-aware features (\cref{tab:feature_ablation}).
Additionally, we also analyze the zero-shot performance of MOHO and the sensitivity of the input hand pose predictions in the \supp.

We first exhibit the effects of the ObMan-based \cite{hasson2019learning} 3D-supervised synthetic-to-real training proposed by \cite{ye2022s} on the top block of \cref{tab:training_ablation}.
Results show that such a strategy enhances the quality of geometric reconstruction (0.12 of F-5 for IHOI on HO3D), even on the larger-scale DexYCB (0.02 of F-5 for gSDF).
Then, we conduct our proposed synthetic-to-real framework on 2D-supervised PixelNeRF to demonstrate its effectiveness.
PixelNeRF directly trained on HO3D performs much inferiorly.
However, after adopting our proposed synthetic-to-real framework, PixelNeRF gains superior results (0.07 on F-5).
The same result is observed on DexYCB for PixelNeRF (0.05 enhancement on F-5).
As for MOHO, the synthetic-to-real framework brings a boost of 0.09 and 0.08 of F-5 on two datasets respectively.
When the amodal-mask-weighted geometric supervision (AMW) is removed, the performance of F-5 decreases by 0.02 and 0.03 on the two datasets respectively.
Moreover, we find that the proposed framework also improves the novel view synthesis results due to suitable knowledge transfer.

\cref{tab:feature_ablation} presents the ablations of the imposed domain-consistent occlusion-aware features.
$\mathcal{F}_s$-16 means extending the PCA dimension of the semantic cues to 16.
We find the performance enhancement is limited (0.01 of F-5).
Thus, $\mathcal{F}_s$-3 setting is used for the MOHO implementation.
Additionally, we compare the global hand-articulated embeddings adopting all hand joints with the local ones adopted in MOHO (\cref{cues}).
The local $\mathcal{F}_h$-L contributes to the performance improvement, since for the specific self-occluded part, the most credible heuristics implied by the holding hand come from the nearest joints.

\section{Conclusion}
This work has presented MOHO for single-view reconstruction of the hand-held object with multi-view occlusion-aware supervision from hand-object videos, tackling two predominant challenges of hand-induced occlusion and object’s self-occlusion.
MOHO presents a novel synthetic-to-real paradigm to unleash hand-induced occlusion by adopting occlusion-free supervisions of SOMVideo in the synthetic pre-training and the amodal-mask-weighted geometric supervision in the real-world finetuning.
Meanwhile, MOHO incorporates domain-consistent occlusion-aware features in order to overcome object’s self-occlusion in the whole synthetic-to-real process.
Extensive experiments on HO3D and DexYCB datasets demonstrate that 2D-supervised MOHO gains superior results against 3D-supervised methods.
In the future, we aim to adopt MOHO for robotic grasping in human-robot interaction scenes.
Limitations are discussed in the \supp. 

\noindent\textbf{Acknowledgement}
This work was supported by the National Key R$\&$D Program of China under Grant 2018AAA0102801.
Moreover, we appreciate Zerui Chen, Yufei Ye and Haowen Sun for kind help.

{\small
\bibliographystyle{ieeenat_fullname}
\bibliography{egbib}
}


\end{document}


\title{Suppelemtary Material for MOHO: Learning Single-view Hand-held Object Reconstruction with Multi-view Occlusion-Aware Supervision}
\author{\authorBlock}
\maketitle

\section{Network Architecture}

\begin{figure*}
    \centering
    \includegraphics[width=0.98\linewidth]{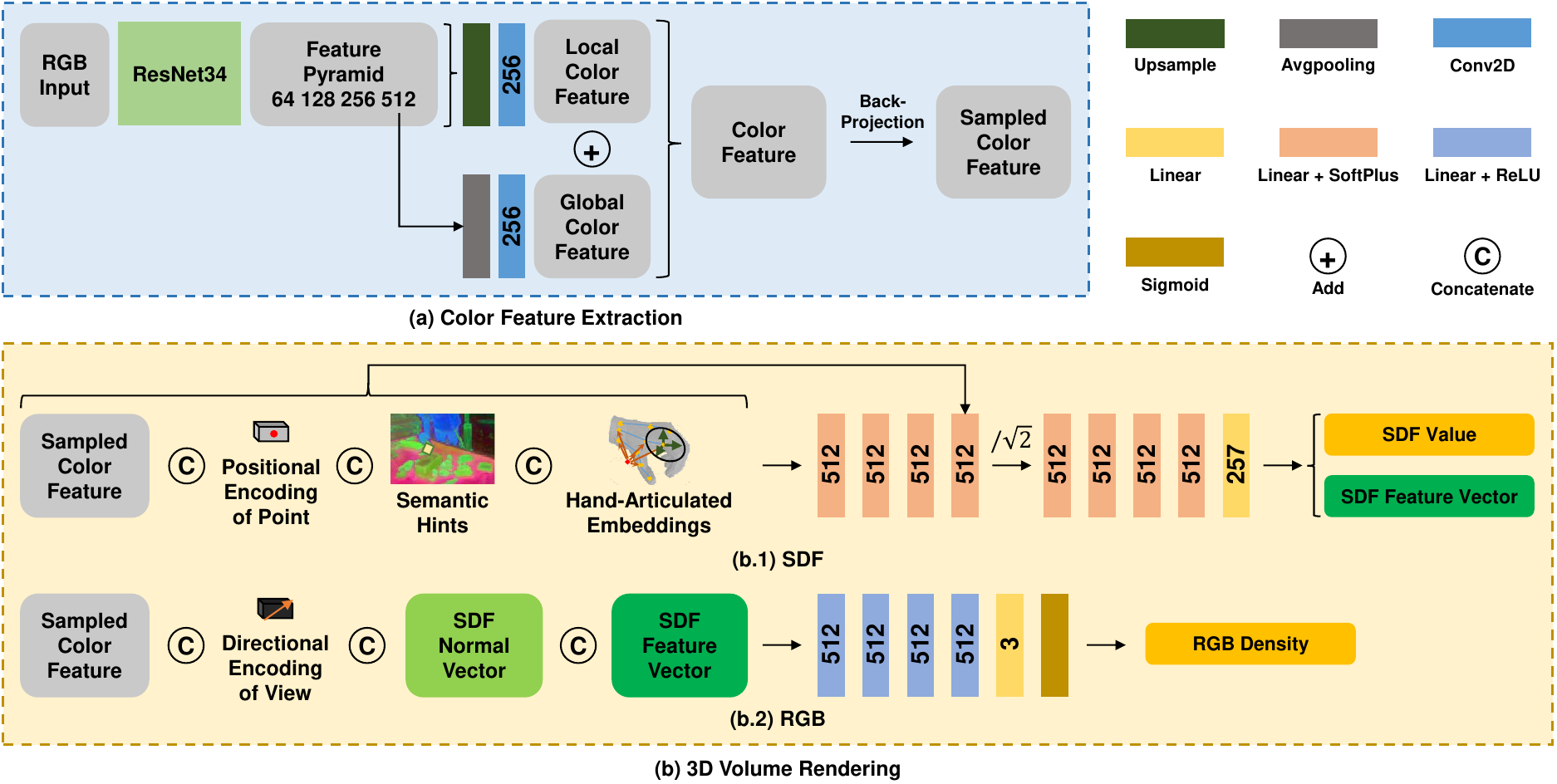}
    \caption{Overview of the MOHO network architecture.}
\label{fig:network}
\end{figure*}

The MOHO network architecture consists of three modules: color feature extraction module, 3D volume rendering head and 2D amodal mask recovery head.
\cref{fig:network} provides an overview of the color feature extraction module and the 3D volume rendering head.

The color feature extraction module bases on ResNet34 \cite{he2016deep}.
We extract feature pyramids using this backbone, and utilize a bottleneck convolutional layer to obtain the local color feature with channel size of 256.
Meanwhile, we use a global average pooling followed by a bottleneck convolutional layer to obtain the global color feature with the same channel size as the local one.
The sum of these two features is back-projected onto the corresponding sampled rays, resulting in the sampled color feature denoted as $\mathcal{F}^i_c$.

For 2D amodal mask recovery head, we utilize a decoder architecture consisting of multi-scale atrous convolution and upsampling network referring to the decoder of DeepLabv3+ \cite{chen2018encoder}, which is applied to obtain probabilistic hand coverage maps by processing the image feature pyramids.

For 3D volume rendering head, we use two MLPs to encode SDF value and RGB density respectively similar to NeuS \cite{wang2021neus}.
The geometric field $\psi_S$ is modeled by an 8-layer MLP with hidden size of 512. 
Softplus with $\beta = 100$ is used as activation function for each hidden layer.
A skip connection with a scale of $\sqrt{2} / 2$ is used at the fourth layer, in order to concatenating the input and intermediate hidden code.
The concatenated point feature $\text{Cat}\left( \mathcal{F}^i_c, E_P\left(\mathcal{P}_i \right), \mathcal{F}^i_s, \mathcal{F}^i_h \right)$ is fed to the geometric field, and a linear layer with output size of 257 is applied at the end to yield a SDF value $s_i$ and a 256-dimensional SDF feature vector $\mathcal{F}^i_{SDF}$ for this sampled point. 
Subsequently, the color field $\psi_C$ is modeled by a 4-layer MLP with ReLU as activation function and hidden size of 512.
The input is the ray feature consisting of $\text{Cat}\left( \mathcal{F}^i_c, E_D\left(\mathcal{D}_i \right), \mathcal{N}^i, \mathcal{F}^i_{SDF} \right)$, where $\mathcal{N}^i$ denotes the normal vector of the geometric field $\mathcal{N}^i = \nabla\psi_S\left(P_i | \mathcal{F}^i_{con} \right)$.
The color field yields 3-dimensional RGB density $c_i$ with the help of a linear layer and a Sigmoid layer.
We apply it to render the color of the pixel by Eq. 4 in the main manuscript.
The $E_P$ and $E_D$ denote the positional and directional encoding functions respectively. 
We apply $E_P$ for spatial location $P_i$ with 6 frequencies and $E_D$ for viewing direction $D_i$ with 4 frequencies.

\section{Details of Synthetic Data Rendering for SOMVideo}

\begin{figure}
    \centering
    \includegraphics[width=0.98\linewidth]{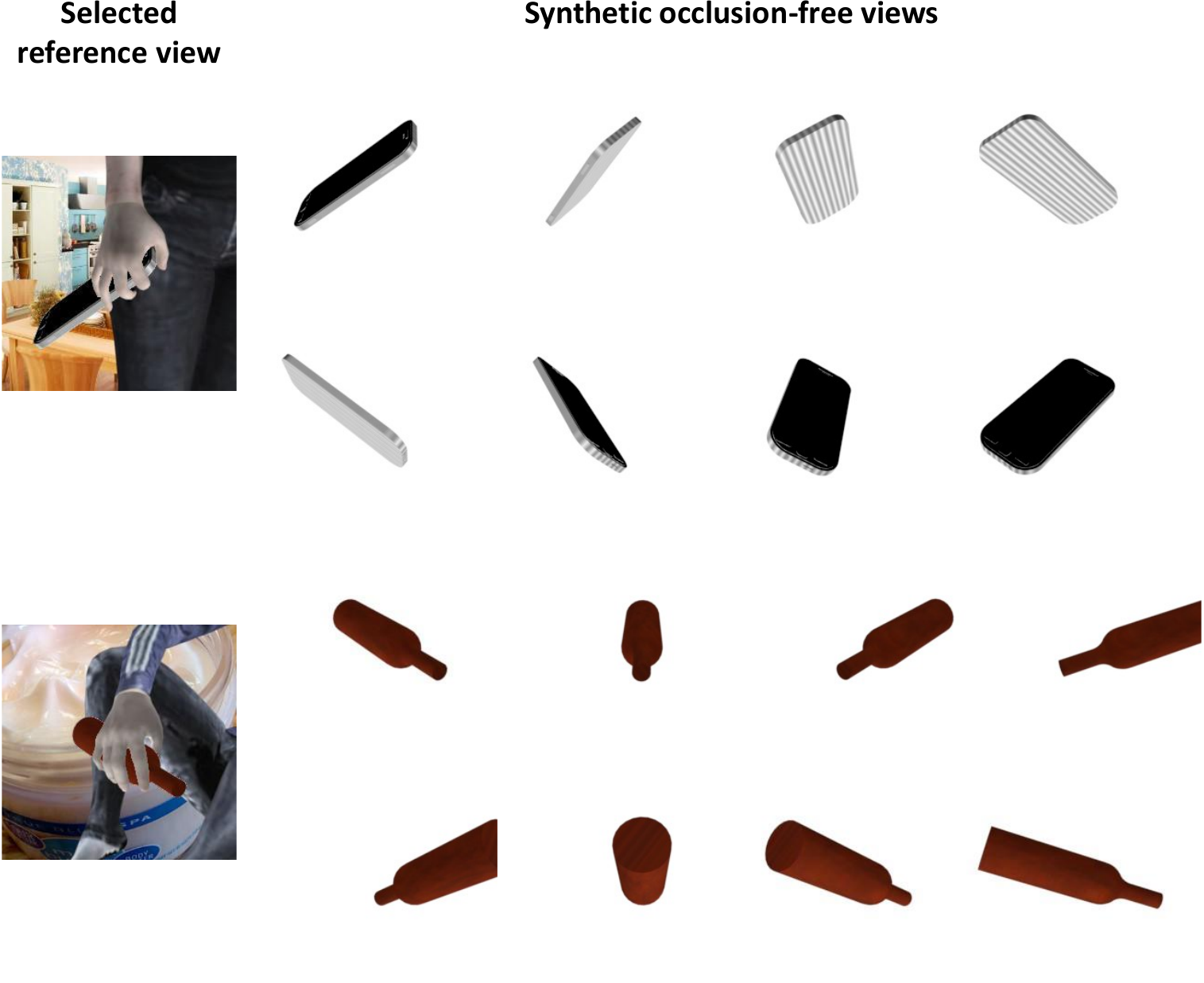}
    \caption{Rendered reference views and occlusion-free views in SOMVideo for our proposed synthetic pre-training.}
\label{fig:render}
\end{figure}

For SOMVideo rendering, we generate each hand-object scene on the basis of the released rendering code of ObMan~\cite{hasson2019learning} dataset. 
Following this setting, we select 8 object categories (bottles, bowls, cans, jars, knifes, cellphones, cameras and remote controls) from ShapeNet \cite{chang2015shapenet} dataset, which results in a total of 2772 meshes.
The object textures are randomly sampled from the texture maps provided with ShapeNet models, and the body textures are sampled from the full body scans used in SURREAL \cite{varol2017learning}.
The skin tone of the hand is matched to the facial color of the body.
The backgrounds are sampled from LSUN \cite{yu2015lsun} and ImageNet \cite{russakovsky2015imagenet} following the ObMan setting.
To render reference views for our synthetic pre-training, we keep the selected shapes, grasps and body poses unchanged as in the ObMan dataset for their plausibility.
Thus, the comparison between our proposed pre-training strategy with the previous 3D-supervised pre-training \cite{ye2022s} adopting ObMan dataset is strictly fair.
We generate 141,550 scenes in total, which exactly corresponds to the scenes in ObMan's training split.
After constructing the hand-object interaction scenes and selecting the reference view, we aim to generate multi-view images capturing such hand-object scenes and occlusion-free supervisions.
To yield them, we fix the position of the grasped object and rotate the camera around it.
The rotated camera trajectory is a circle around the y-axis, centered at the object and with a fixed radius.
The radius is randomly sampled between 50 and 80 cm, kept the same as the implementation of ObMan.
The camera rotates 360 degrees in total, and the video clips are obtained by sampling 10 positions uniformly on the trajectory.
We keep the angle of the camera's rotation around the y-axis equal to the angle of the camera's rotation around its origin, in order to force the camera to focus on the object.
When rendering the corresponding videos without hand-induced occlusion, we only retain the object without the sampled human body in the scene and set the background to white.
Other details are kept exactly the same as the generation process of multi-view hand-object images.
Some examples exhibiting our rendered hand-object reference view and occlusion-free supervising views are shown in Fig. \ref{fig:render}.
The SOMVideo data is released along with our codes.

\section{Additional Loss Terms}
Two additional losses introduced in Sec. 3.3 of the main manuscript regularizing the predicted surface normals are used for restricting the orientation of visible normals towards the camera ($\mathcal{L}_{n_{ori}}$) \cite{verbin2022ref}, and making the predictions smoother ($\mathcal{L}_{n_{smo}}$) \cite{sharma2021point}:
\begin{equation}
\mathcal{L}_{n_{ori}} = \frac{1}{m} \sum_i (min(0, -\hat{n_i}\cdot D_i))^2,
\end{equation}
\begin{equation}
\mathcal{L}_{n_{smo}} = \frac{1}{K} \sum_k (\hat{n_k} - \overline{\hat{n_k}})^2,
\end{equation}
where K is the capacity of K-nearest-neighbor (KNN) region, set to 16 during implementation; $\hat{n_{k/i}} = \sum_j \omega(j)\nabla \psi_S(P(j))$, corresponding to the sampled ray $k$ or $i$; $\overline{\hat{n_k}}$ is the average normal vector in the KNN region.
The definition of $D_i$, $m$, $\omega$, $\psi_S$ and $P$ is kept the same as the main manuscript. 

\section{Limitation Analysis}
\begin{figure}
    \centering
    \includegraphics[width=0.98\linewidth]{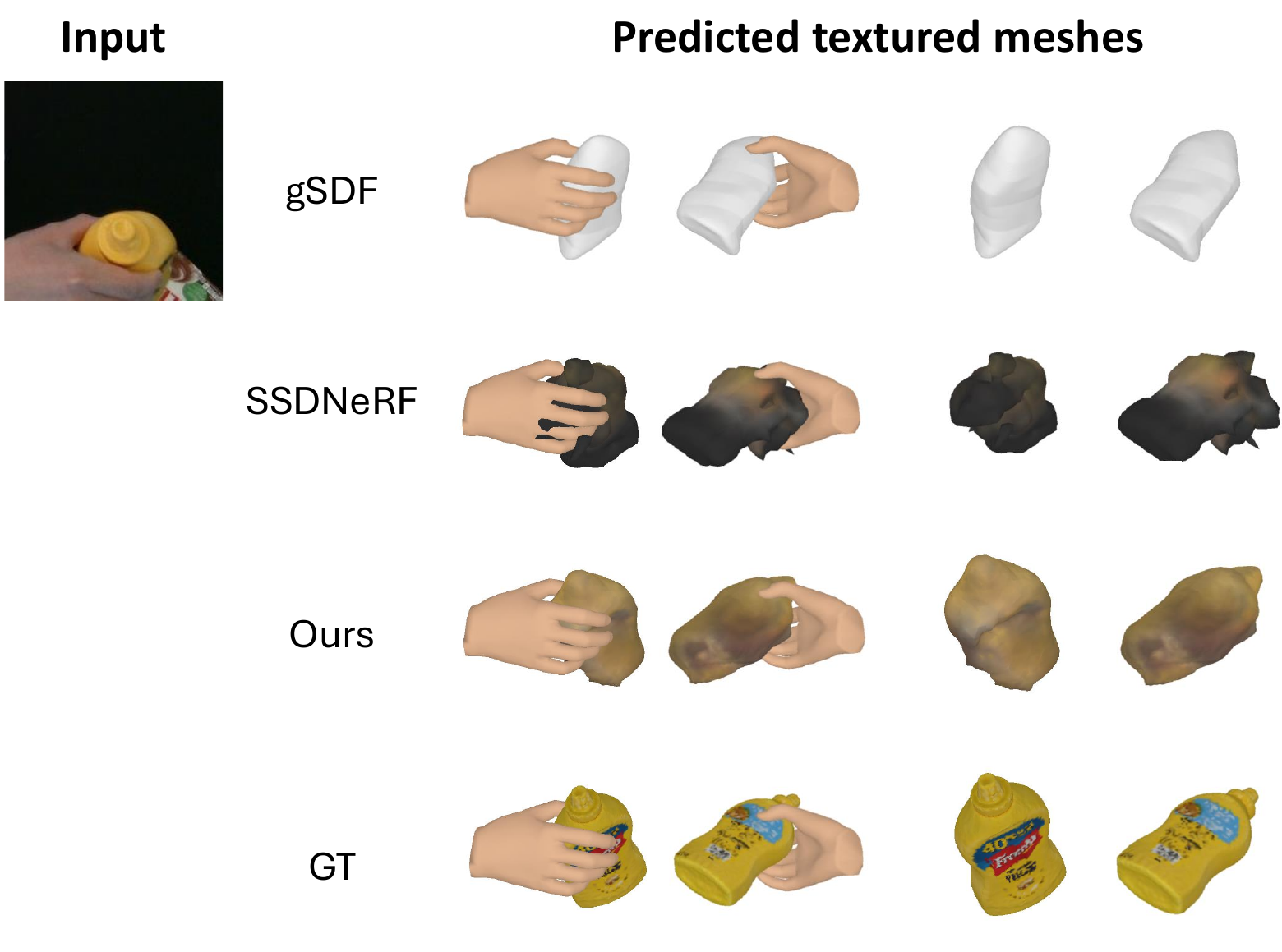}
    \caption{Visualization of failure cases.}
\label{fig:bad_case}
\end{figure}

As shown in Fig. \ref{fig:bad_case}, although MOHO can reconstruct photorealistic textured mesh of hand-held object from a single view, some holes can be found on the reconstructed surface, as well some inconsistent textures are generated.
More advanced backbones or differentiable rendering techniques could be used for better results.
In addition, since current real-world hand-object video datasets are of relatively small scale, the scene, hand and object variety is limited.
The generalization ability across large-scale scene, hand and object variety could be improved for MOHO as new powerful datasets are proposed.

\section{Efficiency Analysis}
To demonstrate the efficiency of MOHO, we compare its running speed to generate the reconstructed object mesh with IHOI, which is the top-performing SDF-based single-view hand-held object reconstruction method.
All experiments are conducted on a single NVIDIA A100 GPU with a reference image as the input (the batch size is set to one).
MOHO runs at 10 FPS, which is slower than IHOI with 23 FPS, but still achieves comparable efficiency. 
The decrement of the inference speed mainly comes from the color branch of our network for texture reconstruction.

\section{Zero-shot Experiments}
\begin{table}[t]
\begin{center}
\tablestyle{2pt}{1.1}
\begin{tabular}{@{}c|ccc|ccc@{}}
\shline
\multirow{2}{*}{Method} & \multicolumn{3}{c|}{HO3D~\cite{hampali2020honnotate}} & \multicolumn{3}{c}{DexYCB~\cite{chao2021dexycb}} \\
& F-5  $\uparrow$ & F-10 $\uparrow$ & CD $\downarrow$ & F-5 $\uparrow$ & F-10 $\uparrow$ & CD $\downarrow$ \\
\hline
IHOI \cite{ye2022s}        &0.14 &0.27  &4.36 &- &-  &- \\
gSDF \cite{chen2023gsdf}        &- &-  &-      & 0.15&0.29  & 1.92\\
Ours        &\textbf{0.23}  &\textbf{0.41}  &\textbf{1.00} & \textbf{0.21}  &\textbf{0.37}  &\textbf{1.24} \\
\shline
\end{tabular}
\end{center}
\caption{Zero-shot experiments of MOHO against 3D-supervised baselines.}
\label{tab:zero_shot}
\end{table}
Tab. \ref{tab:zero_shot} exhibits the zero-shot experiments of MOHO against 3D-supervised baselines.
For fair comparison during implementation, both 3D-supervised baselines IHOI and gSDF are pre-trained on ObMan dataset and directly tested on HO3D and DexYCB respectively.
MOHO is pre-trained on SOMVideo with exactly the same ObMan shapes.
Results show because of the effectiveness of our proposed synthetic pre-training technique for constructing hand-object correlations in both 3D and 2D space, MOHO gains more generalization ability.
Concretely, MOHO exceeds IHOI by 64.2\% of F-5 on HO3D and leads gSDF by 40.0\% of F-5 on DexYCB.

\section{Ablations on the Sensitivity of the Input Hand Pose Predictions}

\begin{table}[t]
\begin{center}
\tablestyle{15pt}{1.1}
\begin{tabular}{@{}c|ccc@{}}
\shline
Noise  &  F-5 $\uparrow$ &  F-10 $\uparrow$ &  CD $\downarrow$ \\
\hline
Pred&0.60  &0.81  &0.15  \\
Pred\,$+\,\sigma$=$0.1$  &0.58  &0.78  &0.16 \\
Pred\,$+\,\sigma$=$0.5$  &0.55  &0.75  &0.18\\
\hline
GT   &0.63  &0.82  &0.14\\
GT\,$+\,\sigma$=$0.1$  &0.60  &0.79  &0.16\\
GT\,$+\,\sigma$=$0.5$  &0.57 &0.76  &0.17 \\
\shline
\end{tabular}
\end{center}
\caption{Ablation studies for the input predicted hand pose on DexYCB~\cite{chao2021dexycb}.}
\label{tab:hand_ablation}
\end{table}

Tab. \ref{tab:hand_ablation} shows the sensitivity of the input hand pose predictions of MOHO.
We add some Gaussian noises with specified variance for this ablation study.
Results illustrate that MOHO gains some robustness against wrong and noisy hand pose predictions.
Meanwhile, if the quality of input hand poses is improved, MOHO yields more accurate reconstruction results, which also demonstrates the effectiveness of our adopted hand-articulated geometric embeddings.

\section{Visual Demonstration of the Occlusion Removal Ability of MOHO}
\begin{figure*}
    \centering
    \includegraphics[width=0.98\linewidth]{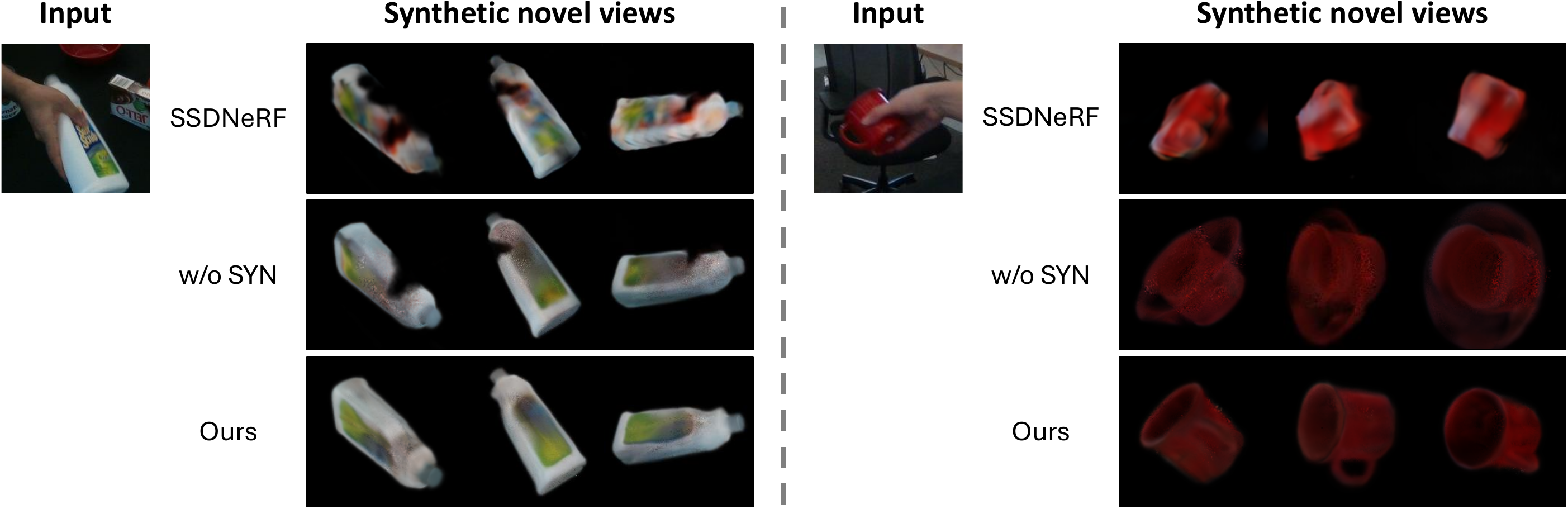}
    \caption{Visual demonstration of the occlusion removal ability.}
\label{fig:novel_view_occlusion}
\end{figure*}

In \cref{fig:novel_view_occlusion}, we compare the visualization results of novel view synthesis to investigate the occlusion removal ability of MOHO.
Specifically, results from SSDNeRF~\cite{chen2023single}, MOHO w/o synthetic pre-training (SYN), and MOHO are exhibited to illustrate the effectiveness of our strategy to resist hand-induced occlusion in real world.

Line 1 indicates that SSDNeRF~\cite{chen2023single} lacks the ability to remove occlusion, which results in the failure to reconstruct hand-covered regions of the input reference view.
The bleach cleanser on the left is reconstructed neglecting the occluded parts (presented as the black fragmentary holes), while the mug on the right is generated with a distorted shape.
The main reason is that the incomplete supervision of real-world videos leads the network only to reconstruct visible parts to get local optimum.
MOHO w/o SYN can get a little more coherent reconstruction though, the occluded parts are still difficult to complete (the bleach cleanser in the left, line 2).
Moreover, the shape distortion is not released utterly due to the lack of complete geometric guidance during training (the mug on the right, line 2).
In contrast, MOHO with the whole synthetic-to-real framework can solve the problem of hand-induced occlusion greatly due to adequate occlusion-aware knowledge transferring.
It generates photorealistic novel views for occluded inputs (Line 3), as well as accurately reconstructs the shape of objects. 

\section{Additional Qualitative Results}

\begin{figure*}
    \centering
    \includegraphics[width=0.98\linewidth]{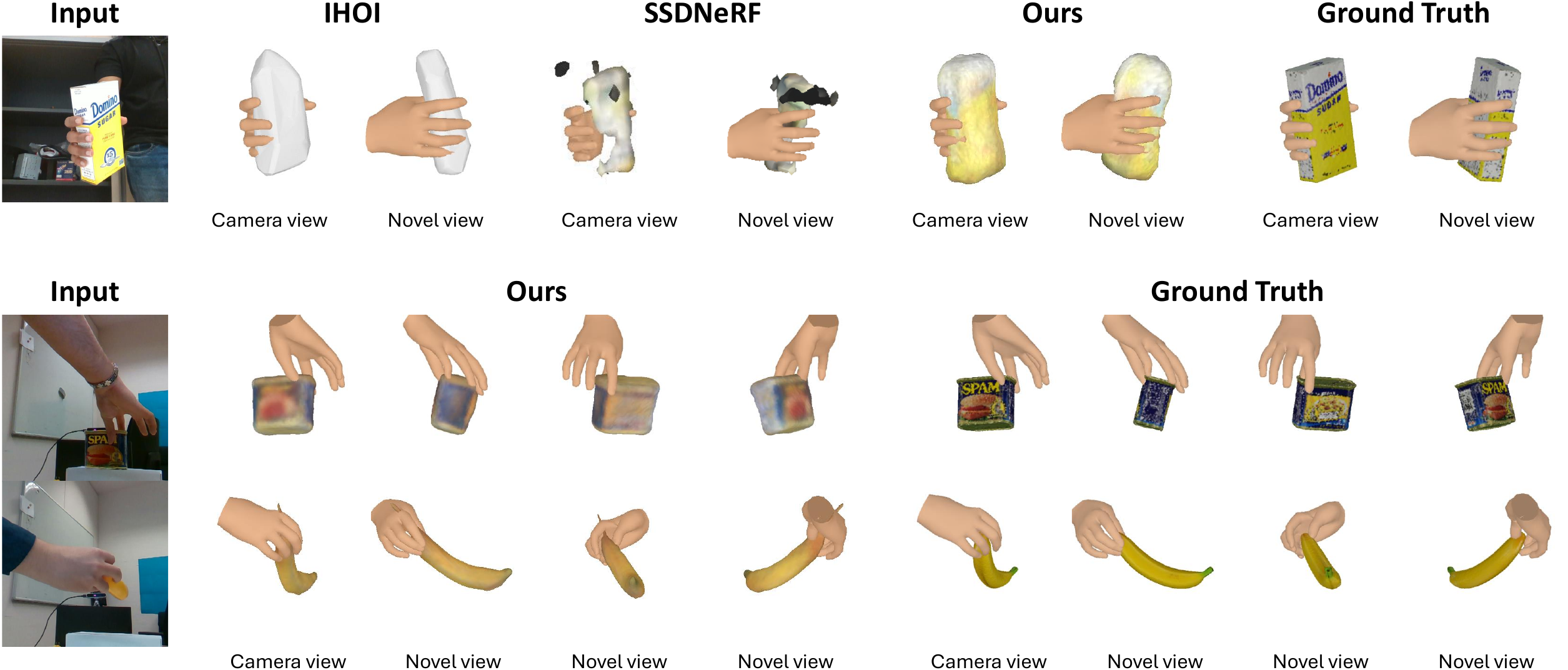}
    \caption{Additional visualization of textured meshes on HO3D~\cite{hampali2020honnotate}.} 
\label{fig:vis_supp_ho3d}
\end{figure*}

\begin{figure*}
    \centering
    \includegraphics[width=0.98\linewidth]{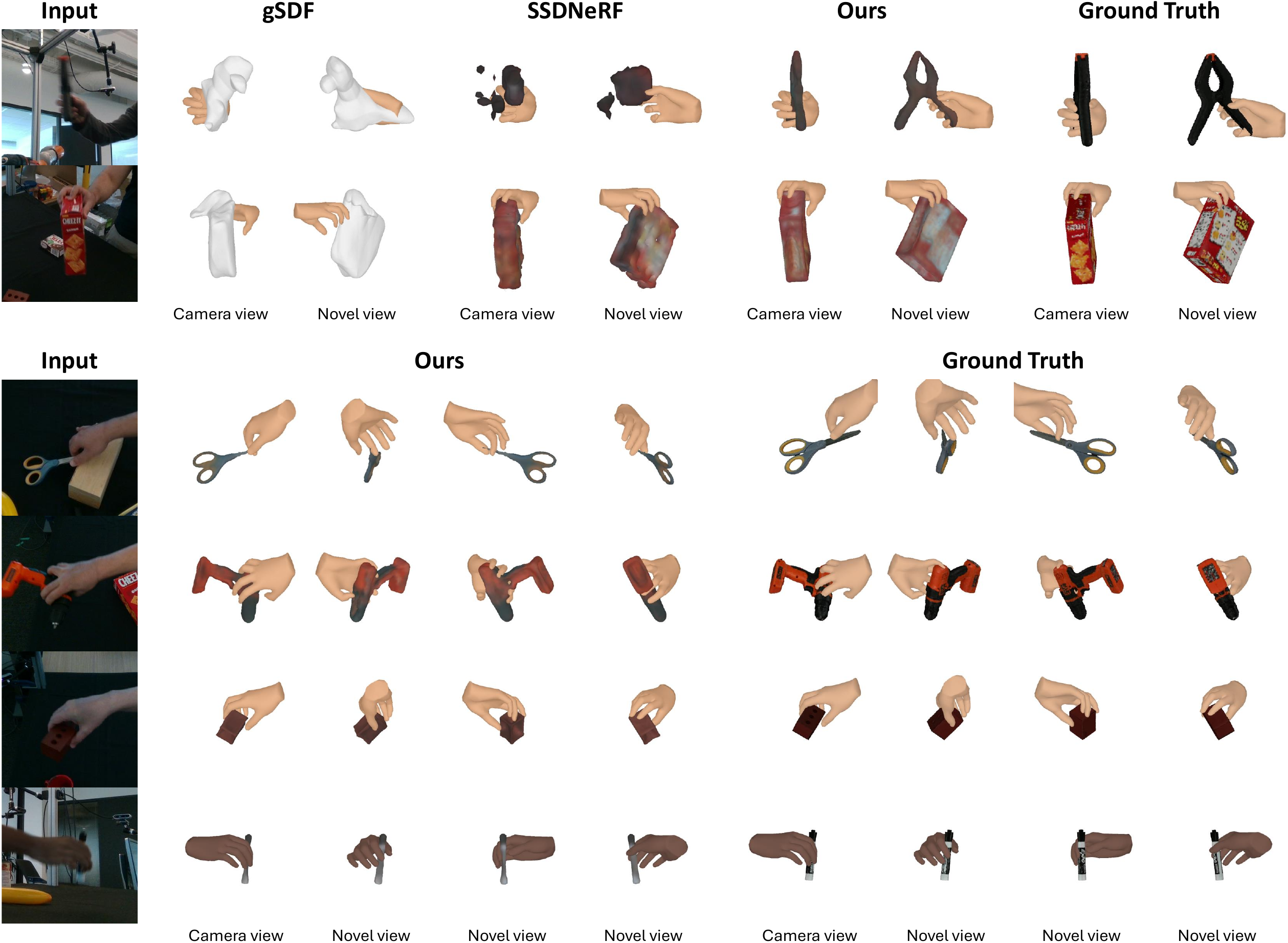}
    \caption{Additional visualization of textured meshes on DexYCB~\cite{chao2021dexycb}.}
\label{fig:vis_supp_dexycb}
\end{figure*}

We visualize additional textured meshes predicted by MOHO and some competitors including IHOI \cite{ye2022s}, gSDF \cite{chen2023gsdf} and SSDNeRF \cite{chen2023single} in \cref{fig:vis_supp_ho3d} and \cref{fig:vis_supp_dexycb} for HO3D \cite{hampali2020honnotate} and DexYCB \cite{chao2021dexycb} respectively.
Compared to the baselines, the predicted textured meshes by MOHO are complete and photorealistic, showing that MOHO releases real-world occlusion obviously and performs well in both mesh reconstruction and texture prediction.

\section{Qualitative Results of Novel View Synthesis}

\begin{figure*}
    \centering
    \includegraphics[width=0.98\linewidth]{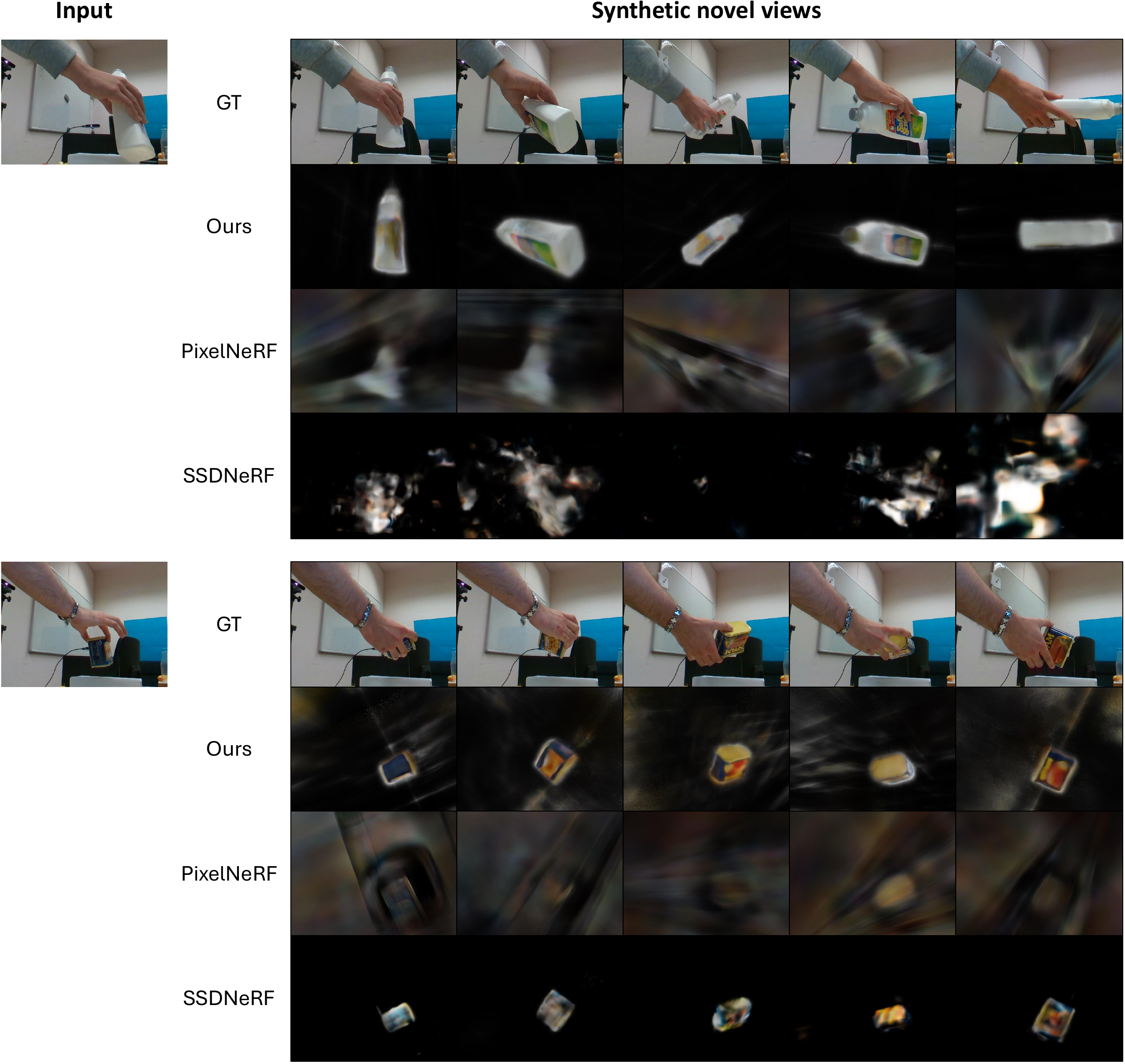}
    \caption{Synthetic novel views on HO3D~\cite{hampali2020honnotate}. }
\label{fig:novel_view_ho3d}
\end{figure*}

\begin{figure*}
    \centering
    \includegraphics[width=0.98\linewidth]{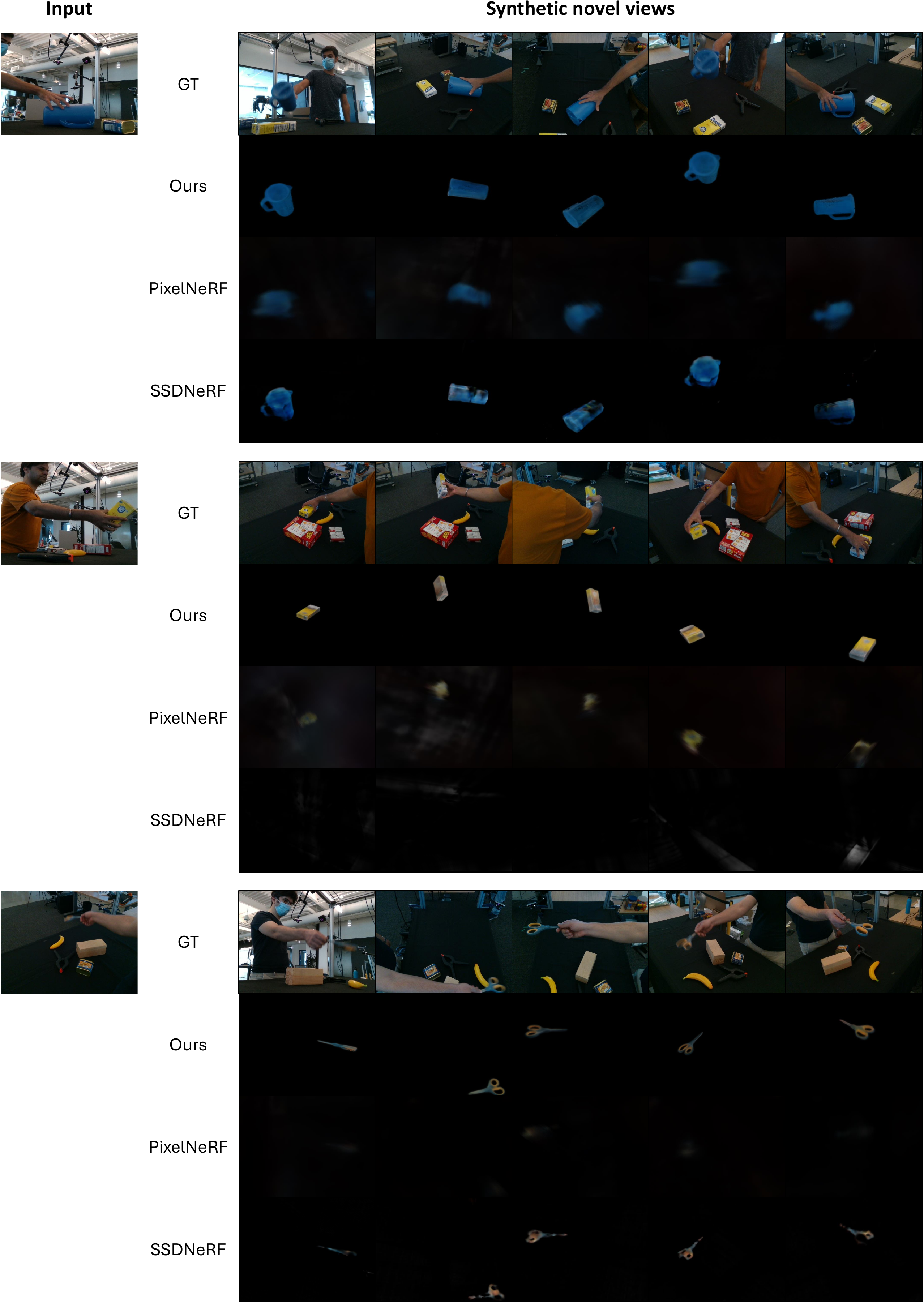}
    \caption{Synthetic novel views on DexYCB~\cite{chao2021dexycb}. }
\label{fig:novel_view_dexycb}
\end{figure*}

We visualize novel view synthesis of MOHO and the NeRF-based competitors PixelNeRF \cite{yu2021pixelnerf} and SSDNeRF \cite{chen2023single} in Fig. \ref{fig:novel_view_ho3d} and \cref{fig:novel_view_dexycb} for HO3D \cite{hampali2020honnotate} and DexYCB \cite{chao2021dexycb} respectively.
Qualitative results on novel view synthesis show due to the imposed partial-to-full cues and the proposed synthetic-to-real framework, MOHO is endowed to handle complex occlusion scenarios in real world and generates more complete, photorealistic, and coherent novel views. 

{\small
\bibliographystyle{ieee_fullname}
\bibliography{egbib}
}